\documentclass{article}
\usepackage{spconf,amsmath,graphicx}
\usepackage{bm}
\usepackage{amsmath}
\usepackage{amssymb}
\usepackage{booktabs}
\usepackage{pifont}  
\usepackage{wasysym} 
\usepackage{multirow}
\usepackage{algorithm}  
\usepackage{algorithmic}

\title{MGRQ: Post-Training Quantization For Vision Transformer \\ With Mixed Granularity Reconstruction}

\name{Lianwei Yang $^{1,2}$, Zhikai Li $^{1,2}$, Junrui Xiao $^{1,2}$, Haisong Gong $^{1,2}$, Qingyi Gu $^{1,}$\sthanks{Corresponding author. \\© 20XX IEEE. Personal use of this material is permitted. Permission from IEEE must be obtained for all other uses, in any current or future media, including reprinting/republishing this material for advertising or promotional purposes, creating new collective works, for resale or redistribution to servers or lists, or reuse of any copyrighted component of this work in other works.}}
  
\address{$^{1}$ Institute of Automation, Chinese Academy of Sciences \\ $^{2}$ School of Artificial Intelligence, University of Chinese Academy of Sciences \\
{\tt\small \{yanglianwei2021, lizhikai2020, xiaojunrui2020, gonghaisong2021, qingyi.gu\}@ia.ac.cn} }

\begin{document}
%
\maketitle
\begin{abstract}
Post-training quantization (PTQ) efficiently compresses vision models, but unfortunately, it accompanies a certain degree of accuracy degradation. Reconstruction methods aim to enhance model performance by narrowing the gap between the quantized model and the full-precision model, often yielding promising results. However, efforts to significantly improve the performance of PTQ through reconstruction in the Vision Transformer (ViT) have shown limited efficacy. In this paper, we conduct a thorough analysis of the reasons for this limited effectiveness and propose MGRQ (\textbf{M}ixed \textbf{G}ranularity \textbf{R}econstruction \textbf{Q}uantization) as a solution to address this issue. Unlike previous reconstruction schemes, MGRQ introduces a mixed granularity reconstruction approach. Specifically, MGRQ enhances the performance of PTQ by introducing Extra-Block Global Supervision and Intra-Block Local Supervision, building upon Optimized Block-wise Reconstruction. Extra-Block Global Supervision considers the relationship between block outputs and the model's output, aiding block-wise reconstruction through global supervision. Meanwhile, Intra-Block Local Supervision reduces generalization errors by aligning the distribution of outputs at each layer within a block. Subsequently, MGRQ is further optimized for reconstruction through Mixed Granularity Loss Fusion. Extensive experiments conducted on various ViT models illustrate the effectiveness of MGRQ. Notably, MGRQ demonstrates robust performance in low-bit quantization, thereby enhancing the practicality of the quantized model.
\end{abstract}

\section{Introduction}
\label{sec:intro}
Recently, Transformer-based visual models have experienced rapid development, showcasing remarkable success in image classification \cite{Crossvit}, object detection \cite{ViT-YOLO}, and instance segmentation \cite{segmentation}. The progress of ViT \cite{Vit} has broadened the horizons of computer vision tasks that were traditionally dominated by convolutional neural network (CNN). However, the computation cost and memory footprint of ViT pose significant challenges compared to CNN, hindering further development and model deployment. Consequently, there is a growing focus on model compression techniques to reduce memory footprint and expedite model inference. Common methods such as distillation \cite{distillation}, pruning \cite{pruning}, and quantization \cite{gholami2022survey} are widely employed in this pursuit.
Quantization converts floating point parameters, such as weights and activations in the model, into low-bit integer parameters. This method is widely employed in model compression due to its benefits of reducing the model's memory cost and accelerating the inference process. Besides, quantization can achieve remarkably high compression ratios without altering the structure.

Quantization inevitably leads to the degradation of model accuracy. Both quantization-aware training (QAT) \cite{choi2018pact,esser2019learned} and post-training quantization (PTQ) \cite{Repq-vit,lin2021fq,liu2021post,yuan2022ptq4vit} aim to address this challenge. PTQ, which acquires quantized models without re-training, offers significant advantages in terms of computational overhead and time efficiency. Although the accuracy of quantized models obtained by PTQ is lower than those from QAT, effective strategies can be employed to narrow this gap. Recent research on PTQ has focused on leveraging reconstruction techniques to enhance the accuracy of quantized models.

\begin{figure}[t]
    \centering
    \includegraphics[width=1\columnwidth]{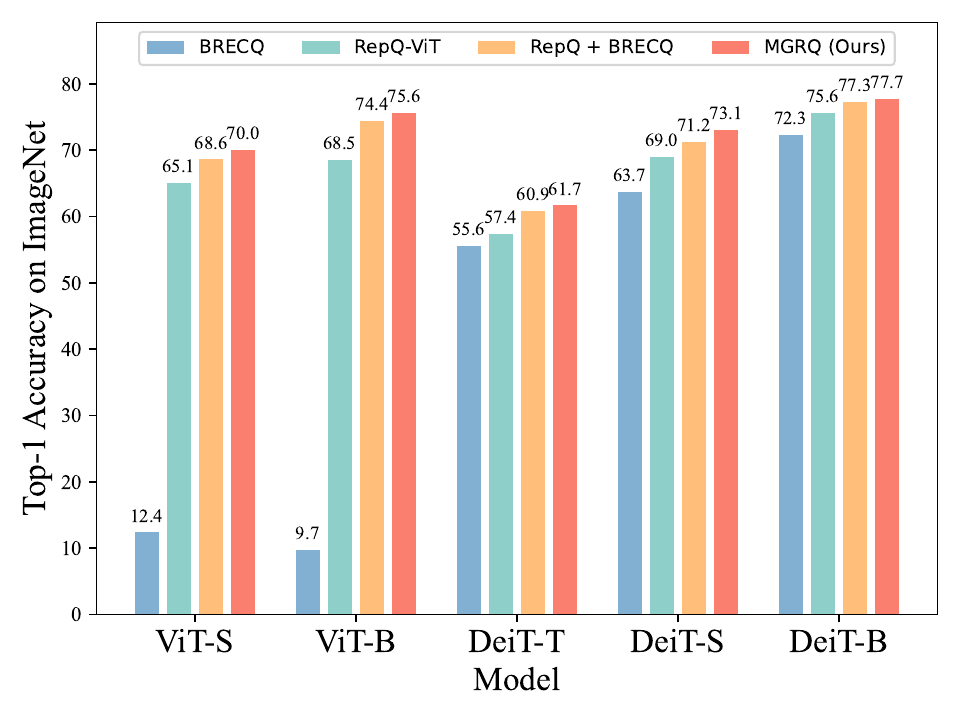}
    \caption{Comparison of Top-1 accuracy on ImageNet after quantizing the model to 4 bits using four different methods.}
\label{fig:ImageNetcompare}
\end{figure}

The reconstruction scheme aims to narrow the performance gap by minimizing the distance between quantized and full-precision models. Brecq \cite{Brecq} exhibits excellent performance gains on CNN, but this is not as evident in ViT. Our analysis suggests that due to the unique structures in ViT, there is a notable decrease in accuracy once quantized. Recognizing this issue, RepQ-ViT \cite{Repq-vit} employs channel-wise quantization and log$\sqrt{2}$ quantization for LayerNorm and Softmax, respectively. This solution significantly mitigates the problem and enhances the performance of 4-bit PTQ to a practical level. However, we do not observe substantial performance improvement when implementing the reconstruction optimization based on RepQ-ViT, as shown in Figure \ref{fig:ImageNetcompare}. This indicates that, in addition to the challenges posed by ViT itself, the reconstruction optimization scheme also encounters some issues. Given that Brecq relies solely on block-wise reconstruction granularity, we posit that this approach falls short in aligning the outputs of each layer to reduce generalization errors. Moreover, it neglects to consider the relationship between block outputs and the final output.

In this paper, our analysis reveals that single-granularity reconstruction is suboptimal, as shown in Figure \ref{fig:ImageNetcompare}. Despite the current success of block-wise reconstruction, it overlooks inter-block dependencies. This oversight is attributed to the inherent limitations of single-granularity reconstruction, which are challenging to overcome independently. Consequently, there is an urgent need to explore mixed granularity reconstruction, where diverse granularity reconstructions mutually complement each other, resulting in superior performance.

Hence, we propose MGRQ, a post-training quantization method for vision transformer with mixed granularity reconstruction. Our method, MGRQ, improves block-wise reconstruction through Extra-Block Global Supervision (EBGS) and Intra-Block Local Supervision (IBLS). EBGS provides global supervision for block reconstruction via the final output, while IBLS aligns layer outputs to minimize generalization errors. The approach concludes with Mixed Granularity Loss Fusion, enhancing overall performance.

In summary, our contributions are as follows:
\begin{itemize}
  \item Our analysis indicates that relying on a single granularity for reconstruction methods is not optimal. While the current mainstream block-wise reconstruction demonstrates commendable performance, its inherent limitations persist, hindering further optimization.
  \item Our approach, MGRQ, skillfully leverages the advantages of different granularities to significantly enhance the model's performance. Our work proposes mixed granularity reconstruction, shifting the initial challenge of selecting the optimal reconstruction granularity into a problem of mixed granularity fusion. 
  \item We have conducted extensive experiments on ImageNet to validate the advantages of MGRQ. It is worth noting that this approach not only enhances the performance of quantized models but also integrates well with methods such as mixed-precision quantization.
\end{itemize}
\section{PRELIMINARIES}
\label{sec:PRELIMINARIES}
\subsection{Vision Transformer Framework}
\label{sssec:Vision Transformer Framework}
Overall, the Vision Transformer consists of a stack of transformer encoder blocks. Each encoder structure primarily consists of multi-head self-attention (MSA) and multi-layer perceptron (MLP). Initially, the ViT transforms the input image into $N$ flattened patches. After the embedding projection, each patch is subsequently mapped to a vector. This tensor is then fed into the encoder block for subsequent operations. Additionally, LayerNorm (LNorm) and residual connections are applied before and after each block, respectively. For a ViT with an $L$-block structure, the realization flow for the $l^{th}$ block is as follows:
\begin{align}
\bm{Y}_{l} &= \bm{X}_{l} + \text{MSA}(\text{LNorm}(\bm{X}_{l})) \\
\bm{X}_{l+1} &= \bm{Y}_{l} + \text{MLP}(\text{LNorm}(\bm{Y}_{l}))
\end{align}
where $l \in \{1, 2, \ldots, L\}$. MSA can capture global correlations between patches. The input to the MSA in the $l^{th}$ block is denoted as $\bm{X}_{l}^{'}$. Following linear projections, we obtain the query, key, and value, respectively.
\begin{align}
\bm{Q}_{l} = \bm{X}_{l}^{'}\bm{W}^q_{i},\quad \bm{K}_{l} = \bm{X}_{l}^{'}\bm{W}^k_{i},\quad \bm{V}_{l} = \bm{X}_{l}^{'}\bm{W}^v_{i}
\end{align}
where $\bm{W_{i}}^{q}$, $\bm{W}_{i}^{k}$, and $\bm{W}_{i}^{v}$ are weight matrices. Subsequently, the attention score is computed through a softmax operation applied to the query and key values. Concatenating the attention scores from $h$ heads results in the output of MSA. 
\begin{align}
\text{Attn}_{i} &= \text{Softmax}\left(\frac{\bm{Q}_{i}\cdot \bm{K}_{i}^T}{\sqrt{D_{h}}}\right)\bm{V}_{i} \\
\text{MSA}(\bm{X}_{l}^{'}) &= [\ \text{Attn}_{1}, \text{Attn}_{2}, ...,\text{Attn}_{h}]\ \bm{W}_o
\end{align}

The MLP aims to project features into a high-dimensional space to capture their representations. This process is achieved through two fully connected layers, incorporating the GELU operation. When the input is $\bm{Y}_{l}^{'}$, the calculation is as follows:
\begin{align}
\text{MLP}(\bm{Y}_{l}^{'}) = \text{GELU}\left(\bm{Y}_{l}^{'}\bm{W}_{1} + \bm{b}_{1} \right)\bm{W}_{2} + \bm{b}_{2}
\end{align}
\subsection{Model Quantization} 
\label{sssec:Model Quantization}
For a pre-trained model with parameters $\theta$, the goal of quantization is to represent $\theta$ using low-bit precision while maintaining high model accuracy. This process is typically executed by a uniform quantizer and operates as follows:
\begin{align}
\label{eq: Initialize model}
\text{Quant: } \bm{x}^{q} &= \text{clip}\left(\left\lfloor\frac{\bm{x}}{s}\right\rceil + z, 0, 2^{b}-1\right) \\
\text{DeQuant: } {\bm{x}^{f}} &= s \cdot (\bm{x}^{q} - z) \approx \bm{x} \\
\text{scaling factor: } {s} &= \frac{\max(\bm{x}) - \min(\bm{x})}{2^b-1} \\  
\text{zero point: } {z} &= \left\lfloor -\frac{\min(\bm{x})}{s} \right\rceil
\end{align}

In Quant, $\bm{x}$ represents the 32-bit floating-point parameter, $b$ indicates the number of bits to be quantized, $\left\lfloor \cdot \right\rceil$ denotes the rounding operation, and the clip function is used to scale the rounded result to the specified range $(0, 2^{b}-1)$, where $\bm{x}^q$ signifies the quantized result. Notably, $s$ is the scaling factor, and $z$ is the zero point, both determined by the boundary value of $\bm{x}$. These parameters are crucial in the calibration operations of PTQ. Dequant recovers the original data from the quantized result. Due to the rounding operation, the outcome may not precisely match the initial full-precision value. 
\section{METHOD}
\label{sec:METHOD}
Reconstructing from block output can yield good results, yet it still neglects the dependencies between blocks. Reconstructing from the network output may seem intuitive, but due to the coarse granularity of the reconstruction, it might not fit the data well. Reconstructing based on layer output serves as a regularization technique, mitigating generalization error by fine-tuning the output distribution of each layer. However, there is a risk of overfitting in the pursuit of approximating the full-precision model. It is evident that reconstruction methods, employing different granularities, encounter technical bottlenecks that are challenging to self-resolve due to their inherent limitations. This motivates us to leverage mixed granularity reconstruction and mitigate issues that may arise from an exclusive reliance on a single granularity reconstruction method.

Our proposed method, MGRQ, consists of three components: Optimized Block-wise Reconstruction (OBWR), Extra-Block Global Supervision (EBGS), and Intra-Block Local Supervision (IBLS). EBGS operates on the basis of block-wise reconstruction, conducting logits matching for the model's final output. Besides, EBGS considers aspects that were overlooked during block reconstruction, enhancing the overall performance. IBLS introduces feature matching within the layers in the block, aligning the block reconstruction process more closely with the full-precision model. The pipeline of MGRQ is summarized in the Algorithm \ref{alg:overview}.
\begin{algorithm}[t]\small
	\caption{Pipeline of MGRQ framework}
	\label{alg:overview}
	\begin{algorithmic}[1]
	\STATE {\bfseries Input:} Pretrained full-precision model \textit{F}, Calibration data $\mathcal{C}$
    \STATE {\bfseries Parameters:} Iteration \textit{T}, Learning rate \textit{lr}, Batchsize \textit{N},  Number of blocks \textit{L}, Hyperparameters $\alpha$ and $\beta$
	\STATE Initialize \textit{F} with $\mathcal{C}$ based on Eq. \ref{eq: Initialize model} to obtain model \textit{Q}
    \FOR{$l=1,2,\cdots,L$}
    \STATE  $\theta^l$ $\leftarrow$ getBlockParameter(\textit{Q}, $l$)
        \FOR{$t=1,2,\cdots,\textit{T}$}
        \STATE $x$ $\leftarrow$ getRandomData($\mathcal{C}$, \textit{N})
        \STATE Obtain ${f}^{f}(x)$ and ${f}^{q}(x)$ based on Eq. \ref{eq:f_logits_out} and Eq. \ref{eq:q_logits_out}
        \STATE Compute $\mathcal{L}_{\scriptscriptstyle EBGS}$ based on Eq. \ref{eq:EBGS}
        \STATE Obtain $m_{l-1}^{f}$ based on Eq. \ref{eq:full-lth-block-out}
        \STATE Compute $\mathcal{L}_{\scriptscriptstyle OBWR}^{\scriptscriptstyle l}$ based on Eq.\ref{eq:OBWR}
        \STATE Obtain ${g}_{i}^{f}$, ${g}_{i}^{q}$ and compute $\mathcal{L}_{\scriptscriptstyle IBLS}^{\scriptscriptstyle l}$ based on Eq.\ref{eq:IBLS}
        \STATE Mixed Granularity Loss Fusion based on Eq. \ref{eq:sumloss}
        \STATE Update the $\theta^{l}$ by backpropagate 
        \ENDFOR
    \ENDFOR
    \STATE {\bfseries Output:} Quantized model $\textit{Q}^{'}$
	\end{algorithmic}
\end{algorithm} 
\subsection{Optimized Block-wise Reconstruction}
\label{ssec: Optimized Block-wise Reconstruction}
Block-wise reconstruction (BWR) is indispensable. The quantized model derived from BWR forms the basis for both Extra-Block Global Supervision (EBGS) and Intra-Block Local Supervision (IBLS), underscoring the integral role of BWR in shaping the final model's performance. Experimental observations in Brecq \cite{Brecq} demonstrate that, when focusing on a single granularity of reconstruction, BWR proves to be the most effective approach. We employ block-wise reconstruction on each transformer encoder block in ViT, and our experiments provide evidence supporting the effectiveness of this configuration. 
We posit that upon completing the reconstruction for the $(l-1)^{th}$ block, the output of the quantized model closely approximates that of the corresponding block in the full-precision model. Consequently, when reconstructing the $l^{th}$ block, favorable reconstruction results are anticipated. We employ the full-precision model as a guide to systematically complete the reconstruction, block by block, starting from the beginning. However, in practical scenarios, the block-wise reconstruction results degrade progressively as the model goes deeper, owing to limitations imposed by the calibration dataset and the accumulation of errors. 

To address this issue, we use the output of the $(l-1)^{th}$ block from the full-precision model as the input for the $l^{th}$ block of the quantized model, resulting in the \textbf{O}ptimized \textbf{B}lock-\textbf{W}ise \textbf{R}econstruction (\textbf{OBWR}). It is crucial to emphasize that our objective is to calculate the loss and does not entail any actual modification to the normal inference process of the quantized model. For the full-precision model and the corresponding quantized model, the role of the $l^{th}$ block is denoted by ${f}_{l}^{f}$ and $f_{l}^{q}$, respectively. The output of the ${(l-1)}^{th}$ block in the full-precision model is denoted by $m_{l-1}^{f}$. As shown in Figure \ref{fig:Extra-Block Global Supervision}, the OBWR for the $l^{th}$ block in the quantized model is computed as follows:
\begin{align}
\label{eq:full-lth-block-out}
m_{l-1}^{f} &= {f}_{l-1}^{f}(m_{l-2}^{f}) \\
\label{eq:OBWR}
\mathcal{L}_{\scriptscriptstyle OBWR}^{\scriptscriptstyle l} &= \|{f}_{l}^{f}(m_{l-1}^{f}) - f_{l}^{q}(m_{l-1}^{f})\|_2
\end{align}

This implies that each block independently approximates the corresponding block of the full-precision model without being influenced by accumulated errors from the previous block. However, this approach neglects the association between the block output and the final actual output of the model. Therefore, we introduce Extra-Block Global Supervision to address this limitation.
\subsection{Extra-Block Global Supervision}
\label{ssec:Extra-Block Global Supervision}
\begin{figure}[t]
    \centering
    \includegraphics[width=1\columnwidth]{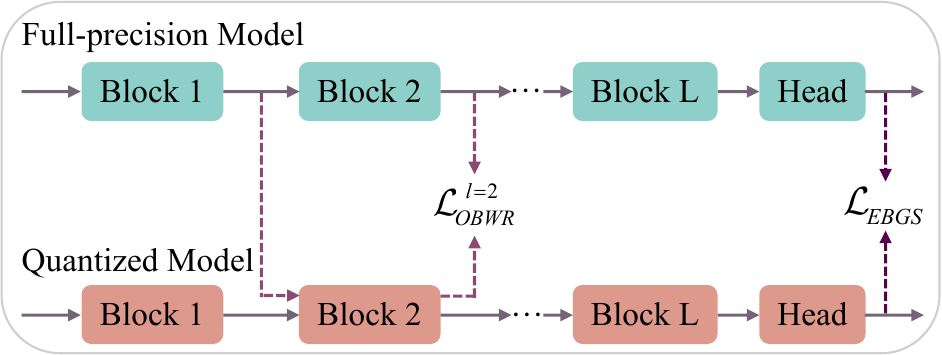}
    \caption{Refining the reconstruction process with Extra-Block Global Supervision on the basis of block-wise reconstruction.}
\label{fig:Extra-Block Global Supervision}
\end{figure}

Given the accumulation of errors arising from block outputs in the actual process, our method MGRQ introduces \textbf{E}xter-\textbf{B}lock \textbf{G}lobal \textbf{S}upervision (\textbf{EBGS}). While the intermediate block output may closely approximate that of the full-precision model, it does not conclusively demonstrate the consistency of the final output. The overall performance of the model is also influenced by the final output. Here, we use ${f}_{l}^{f}$ and $f_{l}^{q}$ to represent the roles of the $l^{th}$ block in the full-precision model and the quantized model, respectively. The same input data, denoted as $x$, is used for both the full-precision and quantized models. As shown in Figure \ref{fig:Extra-Block Global Supervision}, the EBGS in the quantized model is computed as follows:
\begin{align}
\label{eq:f_logits_out}
{{f}^{f}}(x) &= {{f}_{L}^{f}} ({{f}_{L-1}^{f}} ...,  ({{f}_{1}^{f}} ( {x} ) ) ) \\
\label{eq:q_logits_out}
{{f}^{q}}(x) &= {{f}_{L}^{q}} ({{f}_{L-1}^{q}}...,   ({{f}_{1}^{q}} ( {x} ) ) ) \\
\label{eq:EBGS}
\mathcal{L}_{\scriptscriptstyle EBGS} &= \|{{{f}^{f}}(x)}-{{{f}^{q}}(x)}\|_2
\end{align}

The output of each model block is subsequently employed as input for the following block, facilitating continuous inference. The calibration dataset, composed of randomly selected images from the training set, poses challenges in obtaining the corresponding labeling information. Consequently, the Mean Squared Error is directly applied to assess the magnitude of the error between the outputs of the two models.
\begin{table*}[t]
\centering
\small
\begin{tabular}{cccccccc}
\toprule
\textbf{Method}  & \textbf{Opti.} &\textbf{Bit. (W/A)} & \textbf{ViT-S }& \textbf{ViT-B} & \textbf{DeiT-T }& \textbf{DeiT-S} & \textbf{DeiT-B} \\
\midrule
Full-Precision & - & 32/32 & 81.39 & 84.54 & 72.21 & 79.85 & 81.80 \\
\midrule
FQ-ViT \cite{lin2021fq} & $\times$ & 4/4 & 0.10 & 0.10 & 0.10 & 0.10 & 0.10 \\
PTQ4ViT \cite{yuan2022ptq4vit} & $\times$ & 4/4 & 42.57 & 30.69 & 36.96 & 34.08 & 64.39 \\
APQ-ViT \cite{ding2022towards} & $\times$ & 4/4 & 47.95 & 41.41 & 47.94 & 43.55 
 & 67.48 \\
BRECQ \cite{Brecq} & $\checkmark$ & 4/4 & 12.36 & 9.68  & 55.63  & 63.73 & 72.31 \\
QDrop \cite{Qdrop} & $\checkmark$ & 4/4 & 21.24 & 47.30 & 61.93 & 68.27 & 72.60 \\
PD-Quant   & $\checkmark$ & 4/4 & 1.51 & 32.45 & \textbf{62.46} & 71.21 & 73.76 \\
RepQ-ViT \cite{Repq-vit} & $\times$ & 4/4 & 65.05 & 68.48 & 57.43 & 69.03 & 75.61 \\
MGRQ (ours)  & $\checkmark$ & 4/4 & \textbf{70.02} & \textbf{75.59} & 61.71 & \textbf{73.05} &\textbf{77.68} \\
\midrule
FQ-ViT \cite{lin2021fq}   & $\times$ & 6/6 & 4.26 & 0.10 & 58.66 & 45.51 & 64.63 \\
PSAQ-ViT \cite{li2022patch} & $\times$ & 6/6 & 37.19 & 41.52 & 57.58 & 63.61 & 67.95 \\
Ranking-ViT \cite{liu2021post} & $\checkmark$ & 6/6 & - & 75.26 & - & 74.58 & 77.02 \\
EasyQuant  & $\checkmark$ & 6/6 & 75.13 & 81.42 & - & 75.27 & 79.47 \\
PTQ4ViT \cite{yuan2022ptq4vit} & $\times$ & 6/6 & 78.63 & 81.65 & 69.68 & 76.28 & 80.25 \\
APQ-ViT \cite{ding2022towards} & $\times$ & 6/6 & 79.10 & 82.21 & 70.49 & 77.76 & 80.42 \\
NoisyQuant-Linear  & $\times$ & 6/6 & 76.86 & 81.90 & - & 76.37 & 79.77 \\
NoisyQuant-PTQ4ViT  & $\times$ & 6/6 & 78.65 & 82.32 & - & 77.43 & 80.70 \\
BRECQ \cite{Brecq} & $\checkmark$ & 6/6 & 54.51 & 68.33 & 70.28 & 78.46 & 80.85 \\
QDrop \cite{Qdrop} & $\checkmark$ & 6/6 & 70.25 & 75.76 & 70.64 & 77.95 & 80.87 \\
PD-Quant & $\checkmark$ & 6/6 & 70.84 & 75.82 & 70.49 & 78.40 & 80.52 \\
Bit-shrinking \cite{Bit-Shrinking}  & $\checkmark$ & 6/6 & \textbf{80.44} & 83.16 & - & 78.51 & 80.47 \\
RepQ-ViT \cite{Repq-vit}  & $\times$ & 6/6 & 80.43 & 83.62 & 70.76 & 78.90 & 81.27 \\
MGRQ (ours)  & $\checkmark$ & 6/6 & 80.39 & \textbf{83.65} &  \textbf{71.13} & \textbf{79.01} & \textbf{81.38} \\
\bottomrule
\end{tabular}
\vspace{2pt}
\caption{Experimental results on ImageNet for quantized models. ``Opti." indicates optimized models, and ``Bit. (W/A)" signifies W-bit weights and A-bit activation values. In MGRQ, the model is quantized as W4/A4 or W6/A6, except for patch embedding and head, set to W8/A8. Each data shows the Top-1 accuracy (\%) achieved by quantizing the respective model.}
\label{tab:imagenet}
\end{table*}
\subsection{Intra-Block Local Supervision}
\label{ssec:Intra-Block Local Supervision}
\begin{figure}[t]
    \centering
    \includegraphics[width=1\columnwidth]{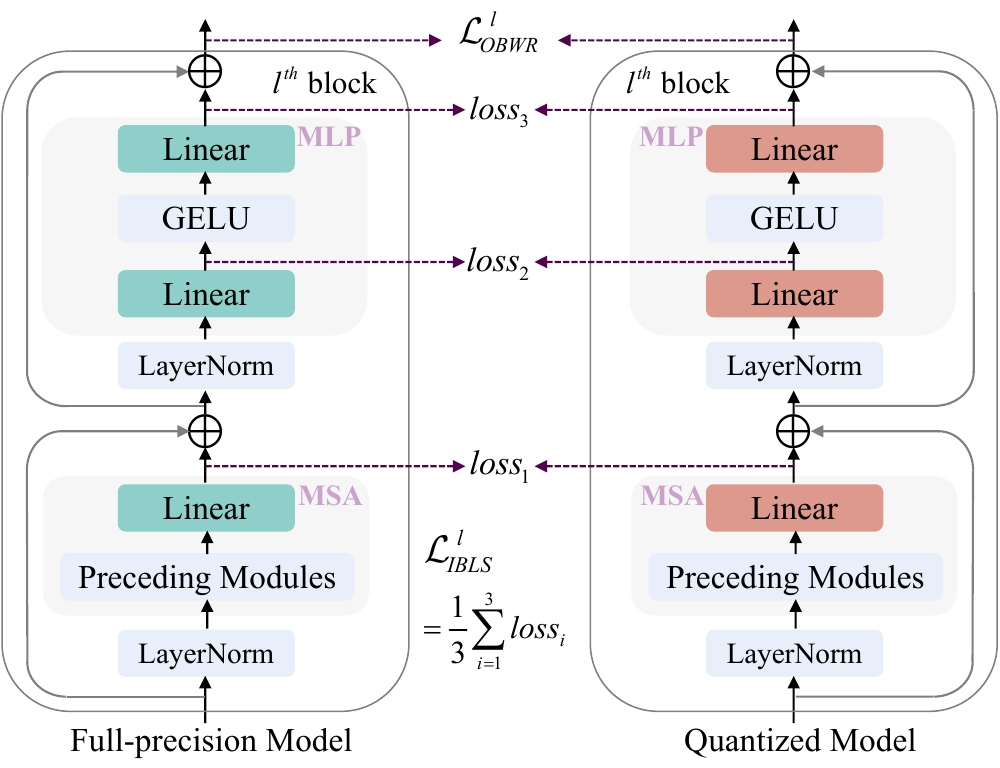}
    \caption{Refining the reconstruction process with Intra-Block Local Supervision on the basis of block-wise reconstruction.}
\label{fig:Intra-Block Local Supervision}
\end{figure}

To enhance accuracy during the block-by-block reconstruction of the output, our method introduces \textbf{I}ntra-\textbf{B}lock \textbf{L}ocal \textbf{S}upervision (\textbf{IBLS}). IBLS focuses on the impact of convolutional or linear layers within the block at a finer granularity, as shown in Figure \ref{fig:Intra-Block Local Supervision}. The evaluation of the disparity between quantized models and full-precision models in IBLS is achieved through feature matching. IBLS, operating at a layer-wise level, acts as a regularizer to minimize generalization error by aligning the output distributions of each layer. 

For the $l^{th}$ block of both the full-precision and quantized models, each consisting of $n$ convolutional or linear layers, the output of each convolutional or linear layer is denoted by $g_{i}^{f}$ and $g_{i}^{q}$, respectively. The computation of IBLS for the $l^{th}$ block is as follows:
\begin{align}
\label{eq:IBLS}
\mathcal{L}_{\scriptscriptstyle IBLS}^{\scriptscriptstyle l} &= \frac{1}{n}\sum_{i=1}^{n} {loss}_{i} =\frac{1}{n}\sum_{i=1}^{n} \|{{g}_{i}^{f}}-{{g}_{i}^{q}}\|_2
\end{align}
\subsection{Mixed Granularity Loss Fusion}
\label{ssec:Loss Trade-off}
In our proposed method, MGRQ, Optimized Block-wise Reconstruction serves as the foundation. Extra-Block Global Supervision enhances block reconstruction by analyzing the correlation between the block output and the final output of the model. Additionally, Intra-Block Local Supervision acts as a regularizer, aligning the output distribution of each layer within the block to mitigate generalization errors. 

Consequently, the total loss requires adjustment based on the scope and significance of different losses. An effective loss fusion strategy directly influences the final reconstruction outcome. We introduce hyperparameters $\alpha$ and $\beta$ to weight the losses and provide the rationale for selecting these values in the experimental setup. Notably, these hyperparameters are also block-wise, reflecting the variations in losses during each phase of the reconstruction. 

As shown in Figure \ref{fig:Extra-Block Global Supervision} and Figure \ref{fig:Intra-Block Local Supervision}, MGRQ performs Mixed Granularity Loss Fusion after obtaining three losses. Simply put, when reconstructing the $l^{th}$ block, the total loss can be formulated as follows:
\begin{align}
\label{eq:sumloss}
\mathcal{L} &= \mathcal{L}_{\scriptscriptstyle OBWR}^{\scriptscriptstyle l} + \alpha \times \mathcal{L}_{\scriptscriptstyle EBGS} + \beta \times \mathcal{L}_{\scriptscriptstyle IBLS}^{\scriptscriptstyle l}
\end{align}
\begin{table}[t]
\centering
\tabcolsep=1mm
\small
\begin{tabular}{cccccc}
\toprule
\textbf{Model} & \textbf{OBWR} &\textbf{EBGS} &\textbf{IBLS} & \textbf{Top-1 (\%)} & \textbf{Bit(W/A)} \\
\midrule
& \multicolumn{3}{c}{Full-Precision} & 79.85 & 32/32 \\
\cmidrule(r){2-6}
 &$\times$&$\times$ & $\times$ & 69.60 & 4/4   \\
 &$\checkmark$ & $\times$ & $\times$  & 71.15  & 4/4 \\
DeiT-Small & $\times$&$\checkmark$ &  $\times$ & 64.26  & 4/4  \\
 & $\times$&$\times$ &$\checkmark$   & 71.28  & 4/4  \\
 &$\checkmark$ &$\checkmark$ & $\times$  & 72.75 & 4/4 \\
 &$\checkmark$ &$\checkmark$ & $\checkmark$ & \textbf{73.05}  & 4/4   \\
\midrule
& \multicolumn{3}{c}{Full-Precision} & 81.39 & 32/32 \\
\cmidrule(r){2-6}
 &$\times$&$\times$ & $\times$ & 65.84 & 4/4   \\
 &$\checkmark$ & $\times$ & $\times$  & 68.58  & 4/4 \\
ViT-Small & $\times$&$\checkmark$ &  $\times$ & 62.15  & 4/4  \\
 & $\times$&$\times$ &$\checkmark$   & 68.91  & 4/4  \\
 &$\checkmark$ &$\checkmark$ & $\times$  & 69.82 & 4/4 \\
 &$\checkmark$ &$\checkmark$ & $\checkmark$ & \textbf{70.02}  & 4/4   \\
 \midrule
& \multicolumn{3}{c}{Full-Precision} & 84.54 & 32/32 \\
\cmidrule(r){2-6}
 &$\times$&$\times$ & $\times$ & 68.57 & 4/4   \\
 &$\checkmark$ & $\times$ & $\times$  & 74.40  & 4/4 \\
ViT-Base & $\times$&$\checkmark$ &  $\times$ & 61.95  & 4/4  \\
 & $\times$&$\times$ &$\checkmark$   & 73.58  & 4/4  \\
 &$\checkmark$ &$\checkmark$ & $\times$  & 75.27 & 4/4 \\
 &$\checkmark$ &$\checkmark$ & $\checkmark$ & \textbf{75.59}  & 4/4   \\
\bottomrule
\end{tabular}
\caption{Ablation study on 4-bit quantized models. We quantize the model to W4/A4, except for patch embedding and head, set to W8/A8. ``OBWR" denotes Optimized Block-wise Reconstruction.``EBGS" denotes Extra-Block Global Supervision. ``IBLS" denotes Intra-Block Local Supervision.}
\label{tab:ablation}
\end{table}
\section{EXPERIMENTS}
\label{sec:EXPERIMENTS}
\subsection{Experimental Setup}
\label{ssec:Experimental setup}
We conduct extensive experiments on various ViT network architectures, including DeiT-Tiny \cite{deit}, DeiT-Small \cite{deit}, DeiT-Base \cite{deit}, ViT-Small \cite{Vit}, and ViT-Base \cite{Vit} models. To evaluate the effectiveness of our proposed method, MGRQ, we compare its performance with that of other methods in terms of accuracy on the ImageNet classification task. All experiments are conducted on a hardware platform with a GeForce A6000 GPU and an AMD EPYC 7272 CPU. The calibration dataset comprises 1024 randomly selected images from the training set of ImageNet. During the reconstruction process, each block undergoes optimization for 3000 iterations using the Adam optimizer with a learning rate set to $1\times10^{-5}$. Optimization is performed for each iteration by randomly selecting 32 images from the calibration dataset. To enhance the quality of reconstruction, our MGRQ method employs a Mixed Granularity Loss Fusion technique. We introduce hyperparameters $\alpha$ and $\beta$ to facilitate loss fusion at different reconstruction granularities. Considering the distinct characteristics of individual ViT blocks, we structure the hyperparameters in a block-wise approach to standardize the different losses to the same scale. ``Wn/An" represents the quantization of both weights and activation values to n bits.
\subsection{Quantization Results on ImageNet Dataset}
\label{ssec:Results on ImageNet}
We conduct experiments, specifically applying W4/A4 and W6/A6 quantization, on various transformer models for image classification using the ImageNet dataset, achieving superior results. Given the advantages of MGRQ, we focus on the performance of low-bit quantization. In W4/A4 quantization, FQ-ViT experiences a significant decrease, reaching only 0.1\%, as shown in Table \ref{tab:imagenet}. While PTQ4ViT and APQ-ViT involve optimization, they do not distinctly enhance accuracy to a practical level. Leveraging the benefits of Extra-Block Global Supervision and Intra-Block Local Supervision, MGRQ demonstrates robust performance at W4/A4.

In W4/A4 quantization, MGRQ demonstrates accuracy improvements of 4.97\%, 7.11\%, 4.02\%, and 2.07\% for ViT-S, ViT-B, DeiT-S, and DeiT-B models, respectively. We conclude that MGRQ plays a crucial role in enhancing the performance of low-bit quantization. MGRQ's accuracy improvement in the classification task is remarkable, narrowing the gap with the full-precision model and enhancing the model's usability after low-bit quantization. Regrettably, we observe that MGRQ does not achieve the most superior results in the DeiT-T model. We attribute this to the limited number of parameters in DeiT-T, which limits the potential of MGRQ.

 In W6/A6 quantization, the MGRQ results, while still favorable, do not exhibit a remarkably significant improvement over W4/A4. We observe that the current method has achieved a satisfactory level of accuracy performance when quantizing the model to 6 bits. The gap between the full-precision model and the 6-bit quantized model obtained by the current mainstream methods is less than 1\%. This somewhat hinders the effectiveness of MGRQ. Nevertheless, we posit that MGRQ has the ability to contribute to narrowing the gap with full-precision models in the future and enhancing the low-bit quantization of the models to more usable levels.
\subsection{Ablation Studies}
To validate the efficacy of the key components in MGRQ, we conduct ablation experiments focusing on Extra-Block Global Supervision (EBGS) and Intra-Block Local Supervision (IBLS). In our extensive ablation experiments involving various transformer models, we observe that the components of MGRQ show significant results in model quantization.

In the ablation study, we quantize the model to 4 bits and use it as a baseline. Subsequently, we introduce Optimized Block-wise Reconstruction (OBWR) and achieve positive results. We conduct separate experimental setups for Extra-Block Global Supervision (EBGS) and Intra-Block Local Supervision (IBLS), respectively, as shown in Table \ref{tab:ablation}. However, when relying solely on network-wise reconstruction like EBGS, the model's performance degrades significantly. We conjecture that this degradation may be attributed to underfitting. Specifically, DeiT-S, ViT-S, and ViT-B experience degradations of 5.34\%, 3.69\%, and 6.62\%, respectively, which are deemed unacceptable. In contrast, utilizing only layer-wise reconstruction, such as IBLS, reveals a performance improvement over the baseline. This suggests the effectiveness of layer-wise reconstruction, but there is a risk of overfitting as the number of training iterations increases.

Building upon Optimized Block-wise Reconstruction, the introduction of Extra-Block Global Supervision (EBGS) leads to accuracy improvements of 1.60\%, 1.24\%, and 0.87\% for DeiT-S, ViT-S, and ViT-B, respectively. This underscores the effectiveness of EBGS in enhancing the accuracy of the model through global supervision. With the subsequent addition of Intra-Block Local Supervision (IBLS), there is an additional improvement of about 0.3\% across different ViT models. This demonstrates that IBLS continues to enhance the performance of the model through local supervision, particularly when the model already exhibits high performance.
\section{CONCLUSIONS}
\label{sec:CONCLUSIONS}
We propose MGRQ, a post-training quantization method designed for ViT with mixed granularity reconstruction. MGRQ initiates with a thorough analysis of the limitations inherent in existing reconstruction schemes for ViT. Subsequently, we propose two key components based on Optimized Block-wise Reconstruction: EBGS and IBLS. EBGS is crafted to consider the relationships between block outputs and the model's output, facilitating global supervision. Meanwhile, IBLS focuses on reducing generalization errors by aligning the output distribution at each layer within a block. Through an extensive array of experiments, we validate the efficacy of MGRQ. The results demonstrate that MGRQ holds significant advantages, especially in low-bit quantization. MGRQ effectively mitigates the issue of model accuracy degradation resulting from post-training quantization. Moreover, it brings low-bit quantization to a practical and usable level.

\vfill\pagebreak
\bibliographystyle{IEEEbib}
\bibliography{strings}

\begin{thebibliography}{10}

\bibitem{Crossvit}
Chun-Fu~Richard Chen, Quanfu Fan, and Rameswar Panda,
\newblock ``Crossvit: Cross-attention multi-scale vision transformer for image classification,''
\newblock in {\em Proceedings of the IEEE/CVF international conference on computer vision}, 2021, pp. 357--366.

\bibitem{ViT-YOLO}
Zixiao Zhang, Xiaoqiang Lu, Guojin Cao, Yuting Yang, Licheng Jiao, and Fang Liu,
\newblock ``Vit-yolo: Transformer-based yolo for object detection,''
\newblock in {\em Proceedings of the IEEE/CVF international conference on computer vision}, 2021, pp. 2799--2808.

\bibitem{segmentation}
Yuqing Wang, Zhaoliang Xu, Xinlong Wang, Chunhua Shen, Baoshan Cheng, Hao Shen, and Huaxia Xia,
\newblock ``End-to-end video instance segmentation with transformers,''
\newblock in {\em Proceedings of the IEEE/CVF conference on computer vision and pattern recognition}, 2021, pp. 8741--8750.

\bibitem{Vit}
Alexey Dosovitskiy, Lucas Beyer, Alexander Kolesnikov, Dirk Weissenborn, Xiaohua Zhai, Thomas Unterthiner, Mostafa Dehghani, Matthias Minderer, Georg Heigold, Sylvain Gelly, et~al.,
\newblock ``An image is worth 16x16 words: Transformers for image recognition at scale,''
\newblock {\em arXiv preprint arXiv:2010.11929}, 2020.

\bibitem{distillation}
Jianping Gou, Baosheng Yu, Stephen~J Maybank, and Dacheng Tao,
\newblock ``Knowledge distillation: A survey,''
\newblock {\em International Journal of Computer Vision}, vol. 129, pp. 1789--1819, 2021.

\bibitem{pruning}
Jinyang Guo, Wanli Ouyang, and Dong Xu,
\newblock ``Multi-dimensional pruning: A unified framework for model compression,''
\newblock in {\em Proceedings of the IEEE/CVF Conference on Computer Vision and Pattern Recognition}, 2020, pp. 1508--1517.

\bibitem{gholami2022survey}
Amir Gholami, Sehoon Kim, Zhen Dong, Zhewei Yao, Michael~W Mahoney, and Kurt Keutzer,
\newblock ``A survey of quantization methods for efficient neural network inference,''
\newblock in {\em Low-Power Computer Vision}, pp. 291--326. Chapman and Hall/CRC, 2022.

\bibitem{choi2018pact}
Jungwook Choi, Zhuo Wang, Swagath Venkataramani, Pierce I-Jen Chuang, Vijayalakshmi Srinivasan, and Kailash Gopalakrishnan,
\newblock ``Pact: Parameterized clipping activation for quantized neural networks,''
\newblock {\em arXiv preprint arXiv:1805.06085}, 2018.

\bibitem{esser2019learned}
Steven~K Esser, Jeffrey~L McKinstry, Deepika Bablani, Rathinakumar Appuswamy, and Dharmendra~S Modha,
\newblock ``Learned step size quantization,''
\newblock {\em arXiv preprint arXiv:1902.08153}, 2019.

\bibitem{Repq-vit}
Zhikai Li, Junrui Xiao, Lianwei Yang, and Qingyi Gu,
\newblock ``Repq-vit: Scale reparameterization for post-training quantization of vision transformers,''
\newblock in {\em Proceedings of the IEEE/CVF International Conference on Computer Vision}, 2023, pp. 17227--17236.

\bibitem{lin2021fq}
Yang Lin, Tianyu Zhang, Peiqin Sun, Zheng Li, and Shuchang Zhou,
\newblock ``Fq-vit: Post-training quantization for fully quantized vision transformer,''
\newblock {\em arXiv preprint arXiv:2111.13824}, 2021.

\bibitem{liu2021post}
Zhenhua Liu, Yunhe Wang, Kai Han, Wei Zhang, Siwei Ma, and Wen Gao,
\newblock ``Post-training quantization for vision transformer,''
\newblock {\em Advances in Neural Information Processing Systems}, vol. 34, pp. 28092--28103, 2021.

\bibitem{yuan2022ptq4vit}
Zhihang Yuan, Chenhao Xue, Yiqi Chen, Qiang Wu, and Guangyu Sun,
\newblock ``Ptq4vit: Post-training quantization for vision transformers with twin uniform quantization,''
\newblock in {\em European Conference on Computer Vision}. Springer, 2022, pp. 191--207.

\bibitem{Brecq}
Yuhang Li, Ruihao Gong, Xu~Tan, Yang Yang, Peng Hu, Qi~Zhang, Fengwei Yu, Wei Wang, and Shi Gu,
\newblock ``Brecq: Pushing the limit of post-training quantization by block reconstruction,''
\newblock {\em arXiv preprint arXiv:2102.05426}, 2021.

\bibitem{ding2022towards}
Yifu Ding, Haotong Qin, Qinghua Yan, Zhenhua Chai, Junjie Liu, Xiaolin Wei, and Xianglong Liu,
\newblock ``Towards accurate post-training quantization for vision transformer,''
\newblock in {\em Proceedings of the 30th ACM International Conference on Multimedia}, 2022, pp. 5380--5388.

\bibitem{Qdrop}
Xiuying Wei, Ruihao Gong, Yuhang Li, Xianglong Liu, and Fengwei Yu,
\newblock ``Qdrop: Randomly dropping quantization for extremely low-bit post-training quantization,''
\newblock {\em arXiv preprint arXiv:2203.05740}, 2022.

\bibitem{li2022patch}
Zhikai Li, Liping Ma, Mengjuan Chen, Junrui Xiao, and Qingyi Gu,
\newblock ``Patch similarity aware data-free quantization for vision transformers,''
\newblock in {\em European Conference on Computer Vision}. Springer, 2022, pp. 154--170.

\bibitem{Bit-Shrinking}
Chen Lin, Bo~Peng, Zheyang Li, Wenming Tan, Ye~Ren, Jun Xiao, and Shiliang Pu,
\newblock ``Bit-shrinking: Limiting instantaneous sharpness for improving post-training quantization,''
\newblock in {\em Proceedings of the IEEE/CVF Conference on Computer Vision and Pattern Recognition}, 2023, pp. 16196--16205.

\bibitem{deit}
Hugo Touvron, Matthieu Cord, Matthijs Douze, Francisco Massa, Alexandre Sablayrolles, and Herv{\'e} J{\'e}gou,
\newblock ``Training data-efficient image transformers \& distillation through attention,''
\newblock in {\em International conference on machine learning}. PMLR, 2021, pp. 10347--10357.

\end{thebibliography}

\end{document}